\documentclass[twoside,11pt]{article}
\usepackage{jmlr2e}

\usepackage[utf8]{inputenc} % allow utf-8 input
\usepackage[T1]{fontenc}    % use 8-bit T1 fonts
\usepackage{hyperref}       % hyperlinks
\usepackage{url}            % simple URL typesetting
\usepackage{booktabs}       % professional-quality tables
\usepackage{amsfonts}       % blackboard math symbols
\usepackage{nicefrac}       % compact symbols for 1/2, etc.
\usepackage{microtype}      % microtypography
\usepackage{times}
\usepackage{graphicx} % more modern
\usepackage{subcaption}
\usepackage{tabularx}
\usepackage{bbm}
\usepackage{url}
\usepackage{amsmath}
\usepackage{bbm}
\usepackage{caption}
\usepackage{float}
\usepackage{amssymb}
\usepackage{grffile}
\usepackage{epstopdf}
\usepackage[colorinlistoftodos]{todonotes}
\usepackage{multirow}

\usepackage{algorithm}
\usepackage{algorithmic}
\usepackage[algo2e,linesnumbered,lined,boxed,vlined]{algorithm2e}

\usepackage{color}

\usepackage[normalem]{ulem}

\newcommand{\hide}[1]{}
\def\b{\ensuremath\boldsymbol}

\usepackage[nobiblatex]{xurl}
\usepackage[numbers]{natbib}

\begin{document}

% \title{Gravitational Dimensionality Reduction: \\Reducing Dimensionality Using Einstein's General Relativity}
% \title{Gravitational Dimensionality Reduction: Metric Learning Using Newtonian Gravity and Einstein's General Relativity}
\title{Gravitational Dimensionality Reduction \\Using Newtonian Gravity and Einstein's General Relativity}

% \author{\name Author(s) \email Email
% }

\author{\name Benyamin Ghojogh \email bghojogh@uwaterloo.ca 
\\
Machine Learning Engineer, Waterloo, ON, Canada
\\
\\
\name Smriti Sharma \email s462shar@uwaterloo.ca \\
David R. Cheriton School of Computer Science, University of Waterloo, ON, Canada
}

% \author{\name Benyamin Ghojogh \email bghojogh@uwaterloo.ca 
% % \\
% % \AND
% % \name Arsalan Sharifnassab \email a.sharifnassab@gmail.com \\
% % \AND
% % \name S.~Jamaloddin Golestani \email golestani@sharif.edu \\
% % \AND
% % \addr 	Department of Sth\\
% % University of Sth \\
% % City
% }
	
%\author{Saber Salehkaleybar,\quad Arsalan Sharifnassab, \quad S.~Jamaloddin Golestani \vspace{1mm} \\ 
%\\
%	\texttt{saleh@sharif.edu},\quad\texttt{a.sharifnassab@gmail.com}, \quad \texttt{golestani@sharif.edu}
%}
\editor{ }

\maketitle

\begin{abstract}

Due to the effectiveness of using machine learning in physics, it has been widely received increased attention in the literature. However, the notion of applying physics in machine learning has not been given much awareness to. This work is a hybrid of physics and machine learning where concepts of physics are used in machine learning. We propose the supervised Gravitational Dimensionality Reduction (GDR) algorithm where the data points of every class are moved to each other for reduction of intra-class variances and better separation of classes. For every data point, the other points are considered to be gravitational particles, such as stars, where the point is attracted to the points of its class by gravity. The data points are first projected onto a spacetime manifold using principal component analysis. We propose two variants of GDR -- one with the Newtonian gravity and one with the Einstein's general relativity. The former uses Newtonian gravity in a straight line between points but the latter moves data points along the geodesics of spacetime manifold. For GDR with relativity gravitation, we use both Schwarzschild and Minkowski metric tensors to cover both general relativity and special relativity. Our simulations show the effectiveness of GDR in discrimination of classes. 
\end{abstract}
\hfill\break

%\tableofcontents

\medskip

\section{Introduction}

% Physics and machine learning can be combined in two ways -- either using machine learning in physics or using physics in machine learning. The former has been noticed more in the literature because machine learning and data science have been found useful in various applications including physics. However, using physics in machine learning is not that much noticed in the literature thus far. 

% Both using machine learning in physics and using physics in machine learning can result in a hybrid of physics and machine learning. Although the former has been addressed in the literature, the latter seems to be more novel. 

There are two ways to combine physics and machine learning -- either using machine learning in physics or using physics in machine learning. The former has been noticed more in the literature because machine learning can be useful in different applications such as physics. However, the latter is more novel. Recently, some works have performed this fusion, focusing more on the hybrid of quantum mechanics and machine learning; they are referred to as quantum machine learning \cite{aimeur2006machine,rupp2015machine,biamonte2017quantum,von2018quantum,teng2018machine,sarma2019machine}.
Physics has also been used in metaheuristic optimization where Newtonian gravity has been used for movement of particles in evolutionary search. That method is called gravitational search algorithm, spanning a family of metaheuristic algorithms \cite{rashedi2009gsa,flores2011gravitational,doraghinejad2014black,rashedi2018comprehensive}.
There is another hybrid work in addition to quantum machine learning and gravitational search algorithm, which models Einstein field equations of general relativity \cite{einstein1916grundlage,lorentz1952principle,wald2010general} in a recurrent neural network \cite{kohli2020einstein}.

Implementing theoretical physics in machine learning can reveal some insights about data patterns. In this paper, we propose Gravitational Dimensionality Reduction (GDR) as a hybrid of physics and machine learning, where physics' concepts are used in machine learning. GDR can be used for dimensionality reduction and metric learning\footnote{Note that (distance) metric learning is a concept in machine learning and dimensionality reduction and it should not be confused with the metric tensor in differential geometry and general relativity.} in machine learning. It is a supervised method and reduces the intra-class variances for better separation of classes of data. It has two variants -- Newtonian gravitational algorithm and relativity gravitational algorithm -- where Newtonian gravity and general relativity's gravity are used respectively for pushing the instances of every class to each other. 
The remaining of this paper is organized as follows. 
Section \ref{section_background_physics} reviews the necessary background on the concepts of physics -- both Newtonian and relativity concepts -- used in this paper.
The proposed GDR algorithm is introduced in Section \ref{section_GDR}, where both Newtonian gravitational algorithm and relativity gravitational algorithm are explained. Simulations are provided in Section \ref{section_simulations} for validation of the proposed algorithm. Finally, Section \ref{section_conclusion} concludes the paper and enumerates the possible future directions. 

\section{Background on Physics}\label{section_background_physics}

\subsection{Newtonian Gravity}

Consider a gravitational particle, such as a planet or a star, with mass $M$. Suppose there is another particle with a smaller mass $m$ with distance $r$ from the particle with mass $M$. According to the Newtonian gravity, developed by Robert Hooke \cite{hooke1665micrographia} and Issac Newton \cite{newton1687principia}, the gravitational force between the two particles is:
\begin{align}\label{equation_Newtonian_gravity}
F = \frac{G M m}{r^2},
\end{align}
where $r$ is the distance of the two particles and $G$ is the gravitational constant, i.e., $G \approx 6.6743 \times 10^{-11}\, m^3 kg^{-1} s^{-2}$.

\subsection{Einstein's General Relativity}

\subsubsection{Einstein's Field Equations}

The Newtonian gravity breaks for high speed movements or high-density masses. Einstein's general relativity \cite{einstein1916grundlage,lorentz1952principle,wald2010general}, proposed by Albert Einstein, generalizes Newtonian gravity to more general cases. Einstein's field equations are ten equations summarized into the following expression:
\begin{align}\label{equation_einstein_field_equations}
G_{\mu \nu} + \Lambda g_{\mu \nu} = \kappa T_{\mu \nu},
\end{align}
where $g_{\mu \nu}$ is the metric tensor of the four-dimensional spacetime manifold, $T_{\mu \nu}$ is the stress-energy tensor modeling all forms of energy (including momentum, flux, stress, and pressure), and $\Lambda \approx 1.1056 \times 10^{-52}\, m^{-2}$ is the cosmological constant (responsible for expansion of universe). 
$\kappa$ is Einstein's gravitational constant:
\begin{align}
\kappa = \frac{8 \pi G}{c^4},
\end{align}
where $G$ is the gravitational constant, introduced before, and $c \approx 299,792,458\, m\, s^{-1}$ is the speed of light in vacuum. 
$G_{\mu \nu}$ is the Einstein tensor, defined as:
\begin{align}\label{equation_einstein_tensor}
G_{\mu \nu} := R_{\mu \nu} - \frac{1}{2} R\, g_{\mu \nu},
\end{align}
where $R_{\mu \nu}$ is the Ricci curvature tensor and $R$ is the scalar curvature. 

Note that $\mu$ and $\nu$ can each take values from $\{0, 1, 2, 3\}$ for the coordinates in the four-dimensional spacetime manifold. By convention, the coordinate $0$ stands for time and coordinates $\{1, 2, 3\}$ are used for the three-dimensional space. This results in $4 \times 4 = 16$ combinations of $\mu$ and $\nu$. However, as the metric tensor is usually chosen to be symmetric, six of these are redundant and, therefore, there are ten Einstein's field equations for the spacetime. 

The left-hand side of Eq. (\ref{equation_einstein_field_equations}), which includes Eq. (\ref{equation_einstein_tensor}) is the curvature of spacetime manifold and its right-hand side, which is the stress-energy tensor, deals with mass and energy. Therefore, Einstein's field equations relate spacetime curvature to mass and energy. Reading Eq. (\ref{equation_einstein_field_equations}) from right to left says that ``a mass tells a spacetime manifold how to curve" and reading it from left to right says that ``a curved spacetime manifold tells a mass how to move"\footnote{This is a famous quote by John Wheeler.}.

\subsubsection{Metric Tensor of Spacetime}

A manifold, such as the spacetime manifold, can be totally described by its metric tensor because a tensor can be used to find the lengths of geodesics (shortest paths) between any two points on the manifold. In other words, a metric tensor is used for generalization of the Pythagorean theorem from flat spaces to curvy manifolds. If $ds$ denotes the infinitesimal distance between two close points on the manifold, we have:
\begin{align}\label{equation_Pythagorean_generalized}
(ds)^2 = \sum_{\mu} \sum_{\nu} g_{\mu \nu}\, dx^\mu\, dx^\nu,
\end{align}
where $dx^\mu$ and $dx^\nu$ are the infinitesimal movement in the directions of $\mu$ and $\nu$ coordinates, respectively, and $g_{\mu \nu}$ is the metric tensor for coordinates $\mu$ and $\nu$.

The Einstein's field equations require differential geometry to be solved \cite{richard2017differential}. They can have various solutions based on the environment and situation of spacetime. Every solution of the Einstein's field equations results in a metric tensor which describes the spacetime manifold. In the following, we introduce two most well-known metric tensors as the solutions of the Einstein's field equations.

% \hfill\break
\textbf{-- Minkowski Metric: }
When there is no mass in the spacetime, meaning that the spacetime is flat, Einstein's general relativity reduces to Einstein's special relativity \cite{einstein1905electrodynamics}. 
In this case, the Minkowski metric \cite{minkowski1908fundamental} is the solution to the Einstein's field equations, where the metric is:
\begin{align}\label{equation_Minkowski_metric}
g = 
\begin{bmatrix}
c^2 & 0 & 0 & 0\\
0 & -1 & 0 & 0\\
0 & 0 & -1 & 0\\
0 & 0 & 0 & -1
\end{bmatrix}.
\end{align}
Note that $g_{\mu \nu}$ denotes one of the elements of the matrix of the metric tensor, depending on the value of $\mu$ and $\nu$.

The three-dimensional space manifold of the Minkowski metric can be considered in the 3D Cartesian coordinate system with coordinates $x$, $y$, and $z$.
This means that the first row and column of metric is for time and the other rows and columns correspond to $x$, $y$, and $z$.

% \hfill\break
\textbf{-- Schwarzschild Metric: }
When there is an uncharged non-rotating gravitational particle (with mass $M$) in the spacetime, the solution to the Einstein's field equations is the Schwarzschild metric \cite{schwarzschild1916gravitational}:
\begin{align}\label{equation_Schwarzschild_metric}
g = 
\begin{bmatrix}
1 - \frac{r_s}{r} & 0 & 0 & 0\\
0 & -(1 - \frac{r_s}{r})^{-1} & 0 & 0\\
0 & 0 & -r^2 & 0\\
0 & 0 & 0 & -r^2 \sin^2\theta
\end{bmatrix},
\end{align}
where $r$ and $\theta$ are the relative distance and the polar angle from the mass in the spherical coordinate system (assuming the mass is at the origin). $r_s$ is the Schwarzschild radius (also called the event horizon of black hole if the mass is a black hole), defined as:
\begin{align}
r_s := \frac{2 G M}{c^2}.
\end{align}

The three-dimensional space manifold of the Schwarzschild metric should be considered in the 3D spherical coordinate system with coordinates $r$ (the radial distance), $\theta \in [0, \pi]$ (the polar angle), and $\phi \in [0, 2\pi]$ (the azimuthal angle).
This means that the first row and column of metric is for time and the other rows and columns correspond to $r$, $\theta$, and $\phi$.
It is noteworthy that the Schwarzschild metric does not include any term with $\phi$ because the mass is assumed to be spherically symmetric and non-rotating. 

\section{Gravitational Dimensionality Reduction}\label{section_GDR}

The goal of the proposed GDR algorithm is to get a $d$-dimensional dataset $\b{D} \in \mathbb{R}^{d \times n}$, having sample size $n$, and transform it to $\b{Y} \in \mathbb{R}^{d \times n}$ in a way that the intra-class variances decrease for better separation of classes\footnote{Note that reducing the intra-class variances is one of the goals of many dimensionality reduction algorithms such as Fisher discriminant analysis \cite{fisher1936use,ghojogh2019fisher}.}. Then, any dimensionality reduction method can be applied to $\b{Y}$ to have a reduced dimensionality of the transformed data. As $\b{Y}$ has better separated classes, it will expected to also have better separated classes after dimensionality reduction.

The GDR algorithm can be viewed in Algorithm \ref{algorithm_GDR}. It has two variants -- with Newtonian and relativity gravitation, shown in Algorithms \ref{algorithm_Newtonian} and \ref{algorithm_relativity}, respectively. As can be seen in Algorithm \ref{algorithm_GDR}, GDR has several steps explained in the following. 

\subsection{Projection onto 3D Space Manifold}

If relativity gravitation is used, we must work in the spacetime manifold because tensor metric is in the spacetime. For simplicity of calculations, we assume that time is frozen so that we work in the 3D space manifold. 
For this, we assume that the 3D space manifold is a 3D PCA subspace. We learn a 3D PCA subspace from the $d$-dimensional $\b{D}$ by solving the following eigenvalue problem \cite{ghojogh2019unsupervised}:
\begin{align}
\b{S} \b{U} = \b{U} \b{\Lambda},
\end{align}
where $\b{S}$ is the covariance matrix of $\b{D}$, $\b{U}$ is the matrix of eigenvectors of $\b{S}$, and $\b{\Lambda}$ is the matrix of eigenvalues of $\b{S}$ \cite{ghojogh2019eigenvalue}.
Truncating the matrix of eigenvectors, to have the vectors with top three eigenvalues, results in $\b{U} \in \mathbb{R}^{d \times 3}$ for the 3D PCA subspace. Projection of data $\b{D}$ onto the column-space of $\b{U}$ gives the data in the 3D space manifold:
\begin{align}
\mathbb{R}^{3 \times n} \ni \b{X} = \b{U}^\top \b{D}.
\end{align}

This projection onto the space manifold is optional for the Newtonian gravitation algorithm because Newtonian gravity can be performed in a Euclidean space with any dimensionality. 

\subsection{Sorting based on Density}

In GDR algorithm, for every data point $\b{x}_j$ within a class we assume that other every data point $\b{x}_i$ within that class is a gravitational particle, such as a planet or a star, which attracts $\b{x}_j$ through gravitation. By this gravitation, $\b{x}_j$ moves in the space. This means that the order of processing matters in the algorithm as movement of a data point $\b{x}_j$ will affect movement of next points. This is because $\b{x}_j$ will also be considered as a gravitational particle for other data points. 

To become robust to the order of processing, we sort the data points of every class based on their density in the class, from largest to smallest density (see line \ref{algorithm_GDR_sort} in Algorithm \ref{algorithm_GDR}). If a point is in a denser region of class, where there are more data points, it is a more important point for representation of class. 
We use the score of Local Outlier Factor (LOF) \cite{breunig2000lof} for this sorting. The more the LOF score of a point, the less the density of that point. Refer to \cite{breunig2000lof} for details of this score. 

After sorting, on one hand, we start taking $\b{x}_j$ from the end of the list (the least important points) because $\b{x}_j$ is going to move and impact movements of next data points. On the other hand, we start taking $\b{x}_i$ from the start of list (the most important points) because $\b{x}_i$ is going to move $\b{x}_j$ so the point $\b{x}_j$ should be first impacted by more important points. 

\subsection{Newtonian Gravitation Algorithm}

In the variant with Newtonian gravitation, the Newtonian gravity, i.e., Eq. (\ref{equation_Newtonian_gravity}), is used. We assume $M=m=G=1$ for simplicity. The distance between $\b{x}_j$ from $\b{x}_i$ is calculated by the Euclidean distance:
\begin{align}\label{equation_r_ij}
r_{ij} = \|\b{x}_i - \b{x}_j\|_2.
\end{align}
We define the amount of movement of $\b{x}_j$ towards $\b{x}_i$ as:
\begin{align}\label{equation_delta_ij}
\delta_{ij} := F \times r_{ij} \overset{(\ref{equation_Newtonian_gravity})}{=} \frac{1}{r_{ij}^2} \times r_{ij} = \frac{1}{r_{ij}},
\end{align}
to have the gravitational force as the fraction of distance $r_{ij}$. 
The movement vector is then calculated as:
\begin{align}
\b{\delta}_{ij} := \delta_{ij} (\b{x}_i - \b{x}_j),
\end{align}
to move $\b{x}_j$ to $\b{x}_i$. $\b{\delta}_{ij}$ is either three-dimensional or $d$-dimensional, depending on whether we are working in the 3D space manifold or the input space of data (recall that PCA projection is optional for the Newtonian gravitation algorithm).
As line \ref{algorithm_GDR_main} in Algorithm \ref{algorithm_Newtonian} demonstrates, we perform this iteratively until convergence. This gradually moves the points of every class towards each other, reducing the intra-class variances. 

\subsection{Relativity Gravitation Algorithm}

In the relativity gravitation variant of GDR, we use Algorithm \ref{algorithm_relativity} in line \ref{algorithm_GDR_main} of Algorithm \ref{algorithm_Newtonian}. 
We use Eq. (\ref{equation_delta_ij}) for calculation of the amount of movement, $\delta_{ij}$; however, we use the metric tensor of space manifold for direction of movement.

\subsubsection{Using Schwarzschild Metric}

If the Schwarzschild metric is used, we should calculate the radial distance and polar angle of $\b{x}_j$ with respect to $\b{x}_i$ in a spherical coordinate system because this metric is in the spherical coordinates. 
We center $\b{x}_j$ using $\b{x}_i$ to put $\b{x}_i$ at the origin of the spherical coordinate system:
\begin{align}
\widehat{\b{x}}_j = \b{x}_j - \b{x}_i.
\end{align}
Then, the radial distance $r_{ij}$ is found by Eq. (\ref{equation_r_ij}). 
Let $\widehat{\b{x}}_j^k$ denote the $k$-th coordinate of $\widehat{\b{x}}_j$.
The polar coordinate $\theta_{ij}$ is obtained by:
\begin{align}
\theta_{ij} = \text{arc}\tan\Big(\frac{\sqrt{(\widehat{\b{x}}_j^1)^2 + (\widehat{\b{x}}_j^2)^2}}{|\widehat{\b{x}}_j^3|}\Big).
\end{align}
If $\widehat{\b{x}}_j^2 < 0$, then the polar angle is adjusted to $\pi - \theta_{ij}$. 

We calculate the space-related elements of the Schwarzschild metric, i.e., Eq. (\ref{equation_Schwarzschild_metric}). Recall that the time was assumed to be frozen for simplicity so only the space-related elements are used. We ignore the negative signs of space-related elements of tensor\footnote{Note that in general relativity, the signs of time-related and space-related elements should be opposite so one can take the time element to be negative and the space elements to be positive.} in Eq. (\ref{equation_Schwarzschild_metric}), but we will encounter the negative signs later. Assuming $G=M=c=1$ for simplicity, the elements of tensor are:
\begin{align}
&g_{rr} = (1 - \frac{2}{r_{ij}})^{-1}, \\
&g_{\theta\theta} = r_{ij}^2, \\
&g_{\phi\phi} = r_{ij}^2 \sin^2(\theta_{ij}).
\end{align}
Noticing that the metric tensor is diagonal, according to Eq. (\ref{equation_Pythagorean_generalized}), we have:
\begin{align}\label{equation_ds_Schwarzschild}
(ds)^2 = g_{rr}\, dr\, dr + g_{\theta\theta}\, d\theta\, d\theta + g_{\phi\phi}\, d\phi\, d\phi,
\end{align}
where $dr$, $d\theta$, and $d\phi$ denote infinitesimal distances along $r$, $\theta$, and $\phi$, respectively. 

We consider the squared movement amount $(\delta_{ij})^2$ to be a linear combination of movements along coordinates $r$, $\theta$, and $\phi$:
\begin{align}\label{equation_delta_alpha_Schwarzschild}
(\delta_{ij})^2 = \alpha_r \delta_{ij} + \alpha_\theta \delta_{ij} + \alpha_\phi \delta_{ij},
\end{align}
where $\alpha_r + \alpha_\theta + \alpha_\phi = 1$ and $\alpha_r, \alpha_\theta, \alpha_\phi \in [0, 1]$.
Setting $\delta_{ij}^2 = (ds)^2$ and comparing Eqs. (\ref{equation_ds_Schwarzschild}) and (\ref{equation_delta_alpha_Schwarzschild}) can give the following:
\begin{align}\label{equation_delta_coordinates_Schwarzschild}
&\delta_{ij}^1 = d_r = -\sqrt{\frac{\alpha_r\, \delta_{ij}}{g_{rr}}}, \quad
\delta_{ij}^2 = d_\theta = -\sqrt{\frac{\alpha_\theta\, \delta_{ij}}{g_{\theta\theta}}}, \quad 
\delta_{ij}^3 = d_\phi = -\sqrt{\frac{\alpha_\phi\, \delta_{ij}}{g_{\phi\phi}}}, 
\end{align}
where we have added back the negative signs that we had dropped before from the tensor metric.
Putting these together gives the movement vector $\b{\delta}_{ij} = [\delta_{ij}^r, \delta_{ij}^\theta, \delta_{ij}^\phi]^\top$ in the spherical coordinate system. 

The centered $\widehat{\b{x}}_j$ is in the Cartesian coordinate system. We need to convert it to the spherical coordinate system to use the movement vector in that coordinate system. We have:
\begin{equation}
\begin{aligned}
&\widehat{\b{x}}_j^r = \sqrt{(\widehat{\b{x}}_j^1)^2 + (\widehat{\b{x}}_j^2)^2 + (\widehat{\b{x}}_j^3)^2}, \\
&\widehat{\b{x}}_j^\theta = \text{arc}\tan\Big(\frac{\sqrt{(\widehat{\b{x}}_j^1)^2 + (\widehat{\b{x}}_j^2)^2}}{\widehat{\b{x}}_j^3}\Big), \\
&\widehat{\b{x}}_j^\phi = \text{arc}\tan\Big(\frac{\widehat{\b{x}}_j^2}{\widehat{\b{x}}_j^1}\Big), 
\end{aligned}
\end{equation}
where we move the data point as:
\begin{align}
&\widehat{\b{x}}_j \gets [\widehat{\b{x}}_j^r, \widehat{\b{x}}_j^\theta, \widehat{\b{x}}_j^\phi]^\top + \b{\delta}_{ij}.
\end{align}
After this movement, we recalculate $r_{ij}$ and $\theta_{ij}$ between $\b{x}_i$ and the moved $\widehat{\b{x}}_j$. The point's representation should be converted back to the Cartesian coordinate system:
\begin{equation}
\begin{aligned}
&\widehat{\b{x}}_j^1 = r_{ij} \sin(\theta_{ij}) \cos(\phi_{ij}), \\
&\widehat{\b{x}}_j^2 = r_{ij} \sin(\theta_{ij}) \sin(\phi_{ij}), \\
&\widehat{\b{x}}_j^1 = r_{ij} \cos(\theta_{ij}).
\end{aligned}
\end{equation}
We add $\b{x}_i$ back to the centered moved point $\widehat{\b{x}}_j$ to have $\b{x}_j = \widehat{\b{x}}_j + \b{x}_i$.
This is performed iteratively until convergence. This moves the points of every class to each other to reduce the intra-class variances. 

\subsubsection{Using Minkowski Metric}

If we use the Minkowski metric, i.e., Eq. (\ref{equation_Minkowski_metric}), the space-related elements of tensor, with ignorance of the negative signs, are:
\begin{align}
g_{xx} = g_{yy} = g_{zz} = 1.
\end{align}
We have:
\begin{align}
&(ds)^2 = (\delta_{ij})^2 = g_{xx}\, dx\, dx + g_{yy}\, dy\, dy + g_{zz}\, dz\, dz, \\
&(\delta_{ij})^2 = \alpha_x \delta_{ij} + \alpha_y \delta_{ij} + \alpha_z \delta_{ij},
\end{align}
where $\alpha_x + \alpha_y + \alpha_z = 1$ and $\alpha_x, \alpha_y, \alpha_z  \in [0, 1]$.
Comparing these and encountering back the negative signs give:
\begin{align}
&\delta_{ij}^1 = d_x = -\sqrt{\frac{\alpha_x\, \delta_{ij}}{g_{xx}}}, \quad
\delta_{ij}^2 = d_y = -\sqrt{\frac{\alpha_y\, \delta_{ij}}{g_{yy}}}, \quad 
\delta_{ij}^3 = d_z = -\sqrt{\frac{\alpha_z\, \delta_{ij}}{g_{zz}}}.
\end{align}
Putting these together gives the movement vector $\b{\delta}_{ij} = [\delta_{ij}^x, \delta_{ij}^y, \delta_{ij}^z]^\top$ in the Cartesian coordinate system. 
The movement of the point $\b{x}_j$ is:
\begin{align}
\b{x}_j \gets \b{x}_j + \b{\delta}_{ij}
\end{align}

\subsection{Final Steps}

Recall that the data points were sorted based on their density.
After the iterative movements of data points, we unsort the data points to their original positions in the dataset to have the transformed $\b{X}$ finally. In this transformed data, the variance of every class has become much smaller so the classes have become more separated. If we were working in the 3D PCA subspace (i.e., the space manifold), we reconstruct the data to the $d$-dimensional input space by $\mathbb{R}^{d \times n} \ni \b{Y} = \b{U} \b{X}$. If we were working in the $d$-dimensional space in the Newtonian gravitational algorithm, we simply get $\b{X}$ as the transformed data $\b{Y}$. 

The transformed data $\b{Y}$ in the $d$-dimensional input space has better separation of classes compared to $\b{D}$. 
If dimensionality reduction is required, it is possible to apply any other dimensionality reduction algorithm to the transformed data $\b{Y}$ to obtain features with less dimensionality and more discrimination of classes in the subspace. 

\SetAlCapSkip{0.5em}
\IncMargin{0.8em}
\begin{algorithm2e}[!t]
\DontPrintSemicolon
    \textbf{Procedure}: GDR()\;
    \textbf{Input}: Training data $\b{D}$\;
    // Project onto the space manifold:\;
    \uIf{work in space manifold}{
        $\b{X} = \b{U}^\top \b{D}$\;
    }
    \Else{
        $\b{X} = \b{D}$\;
    }
    // Sort points based on density:\; \label{algorithm_GDR_sort}
    \For{$k$ from $1$ to $\ell$}{ 
        $\b{X}_k \gets$ Extract instances of class $k$\;
        $\b{X}_k \gets$ Sort($\b{X}_k$)\;
    }
    // Main iterative algorithm:\; \label{algorithm_GDR_main}
    \While{not converged}{
        \For{$k$ from $1$ to $\ell$}{
            $\b{X}_k \gets$ Newtonian($\b{X}_k$) or Relativity($\b{X}_k$)\;
        }
    }
    // Unsort points to original order:\;
    \For{$k$ from $1$ to $\ell$}{
        $\b{X}_k \gets$ Unsort($\b{X}_k$)\;
    }
    $\b{X} \gets$ Put together instances of classes\;
    // Reconstruct from the space manifold:\;
    \uIf{work in space manifold}{
        $\b{Y} = \b{U} \b{X}$\;
    }
    \Else{
        $\b{Y} = \b{X}$\;
    }
    \textbf{Return} $\b{Y}$\;
\caption{GDR algorithm with Newtonian or relativity methods}\label{algorithm_GDR}
\end{algorithm2e}
\DecMargin{0.8em}

\SetAlCapSkip{0.5em}
\IncMargin{0.8em}
\begin{algorithm2e}[!t]
\DontPrintSemicolon
    \textbf{Procedure}: Newtonian()\;
    \textbf{Input}: Training data $\b{X}$ (three-dimensional or $d$-dimensional)\;
    $n \gets$ cardinality of $\b{X}$\;
    \For{$j$ from $n$ to $1$}{
        $\b{\delta}_j \gets \b{0}$\;
        \For{$i$ from $1$ to $n$, where $i \neq j$}{
            $r_{ij} \gets \|\b{x}_i - \b{x}_j\|_2$\;
            $\delta_{ij} \gets \frac{1}{r_{ij}}$\;
            $\b{\delta}_{ij} \gets \delta_{ij} (\b{x}_i - \b{x}_j)$\;
            $\b{\delta}_j \gets \b{\delta}_j + \b{\delta}_{ij}$\;
        }
        $\b{x}_j \gets \b{x}_j + \b{\delta}_j$\;
    }
    \textbf{Return} $\b{X}$\;
\caption{Newtonian gravitation algorithm}\label{algorithm_Newtonian}
\end{algorithm2e}
\DecMargin{0.8em}

\SetAlCapSkip{0.5em}
\IncMargin{0.8em}
\begin{algorithm2e}[!t]
\DontPrintSemicolon
    \textbf{Procedure}: Relativity()\;
    \textbf{Input}: Training data $\b{X}$ (three-dimensional), metric tensor $\b{g}$\;
    $n \gets$ cardinality of $\b{X}$\;
    \For{$j$ from $n$ to $1$}{
        \For{$i$ from $1$ to $n$, where $i \neq j$}{
            $r_{ij} \gets \|\b{x}_i - \b{x}_j\|_2$\;
            $\delta_{ij} \gets \frac{1}{r_{ij}}$\;
            // Metric tensor:\;
            \uIf{Schwarzschild metric}{
                $\widehat{\b{x}}_j \gets \b{x}_j - \b{x}_i$\;
                $\theta_{ij} \gets \text{arc}\tan\Big(\frac{\sqrt{(\widehat{\b{x}}_j^1)^2 + (\widehat{\b{x}}_j^2)^2}}{|\widehat{\b{x}}_j^3|}\Big)$\;
                \If{$\widehat{\b{x}}_j^2 < 0$}{
                    $\theta_{ij} \gets \pi - \theta_{ij}$\;
                }
                $g_{11}, g_{22}, g_{33} \gets (1 - \frac{2}{r_{ij}})^{-1}, r_{ij}^2, r_{ij}^2 \sin^2(\theta_{ij})$\;
            }
            \ElseIf{Minkowski metric}{
                $g_{11}, g_{22}, g_{33} \gets 1, 1, 1$\;
            }
            // Movement:\;
            $\delta_{ij}^1 \gets - \sqrt{\frac{\alpha_1\, \delta_{ij}}{g_{11}}}$\;
            $\delta_{ij}^2 \gets - \sqrt{\frac{\alpha_2\, \delta_{ij}}{g_{22}}}$\;
            $\delta_{ij}^3 \gets - \sqrt{\frac{\alpha_3\, \delta_{ij}}{g_{33}}}$\;
            $\b{\delta}_{ij} \gets [\delta_{ij}^1, \delta_{ij}^2, \delta_{ij}^3]^\top$\;
            
            \uIf{Schwarzschild metric}{
                $\widehat{\b{x}}_j \gets$ Convert $\widehat{\b{x}}_j$ from Cartesian to spherical\;
                $\widehat{\b{x}}_j \gets \widehat{\b{x}}_j + \b{\delta}_{ij}$\;
                $r_{ij} \gets \|\widehat{\b{x}}_j\|_2$\;
                $\theta_{ij} \gets \text{arc}\tan\Big(\frac{\sqrt{(\widehat{\b{x}}_j^1)^2 + (\widehat{\b{x}}_j^2)^2}}{|\widehat{\b{x}}_j^3|}\Big)$\;
                $\widehat{\b{x}}_j \gets$ Convert $\widehat{\b{x}}_j$ from spherical to Cartesian\;
                $\b{x}_j \gets \widehat{\b{x}}_j + \b{x}_i$\;
            }
            \ElseIf{Minkowski metric}{
                $\b{x}_j \gets \b{x}_j + \b{\delta}_{ij}$\;
            }
        }
    }
    \textbf{Return} $\b{X}$\;
\caption{Relativity gravitation algorithm}\label{algorithm_relativity}
\end{algorithm2e}
\DecMargin{0.8em}

\section{Simulations}\label{section_simulations}

For verification of the GDR algorithm, we performed experiments on the sklearn's NIST digit dataset with ten classes of digits, sample size 1797, and dimensionality 64. 
The code of the GDR algorithm can be found in \url{https://github.com/bghojogh/Gravitational-Dimensionality-Reduction}.

\subsection{Verification of GDR algorithm}

Figure \ref{figure_experiments1_Newtonian_and_Relativity} displays the data $\b{X}$ in the 3D PCA subspace (i.e., the space manifold). The data in the PCA subspace are shown for six iterations of the GDR algorithm with Newtonian and relativity gravitation approaches. 
In the experiments of Fig. \ref{figure_experiments1_Newtonian_and_Relativity}, the Schwarzschild metric and $\alpha_r = \alpha_\theta = \alpha_\phi=0.33$ were used for the relativity gravitation approach.
As the figure shows, the classes have been separated because of reduction in their variances in both Newtonian and relativity approaches. 

Moreover, we show the two-dimensional t-SNE representation \cite{van2008visualizing} of the $d$-dimensional data in this figure. At the end of every iteration, we also reconstructed data from the PCA subspace to the $d$-dimensional space and showed the t-SNE representation of modified data at every iteration. As the t-SNE representations show in Fig. \ref{figure_experiments1_Newtonian_and_Relativity}, the classes of data have become more separated after each iteration. 

\subsection{Newtonian Approach in the Input Space}

Figure \ref{figure_experiments1_Newtonian_and_Relativity} also depicts the GDR performance with Newtonian gravitation where data are not projected onto the PCA subspace. It is seen that the Newtonian method also works well in the $d$-dimensional input space. 

\subsection{Comparison of Newtonian and Relativity Approaches}

\begin{figure*}[!t]
\centering
\includegraphics[width=\textwidth]{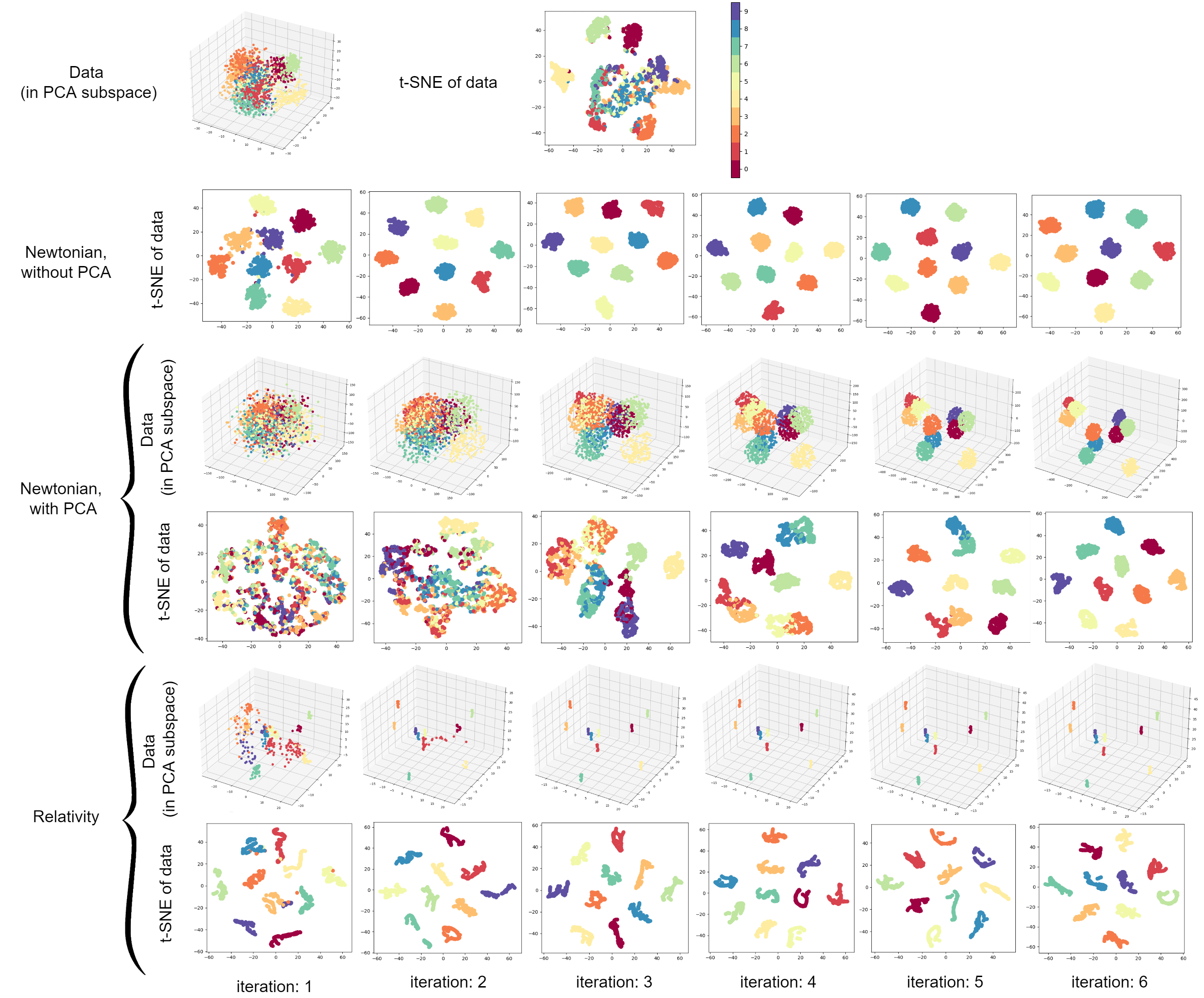}
\caption{Performance of GDR algorithm with Newtonian and relativity gravitation.}
\label{figure_experiments1_Newtonian_and_Relativity}
\end{figure*}

The performances of the Newtonian and relativity gravitation algorithms are also compared in Fig. \ref{figure_experiments1_Newtonian_and_Relativity}. This comparison shows that the relativity approach moves the data points much faster to one another so it is a faster approach in terms of number of iterations. This makes sense because the relativity approach uses the geodesics (shortest paths on the manifold), rather than the straight movements, to move in the space.

\subsection{GDR with Relativity Gravitation Using the Minkowski Metric}

\begin{figure*}[!t]
\centering
\includegraphics[width=\textwidth]{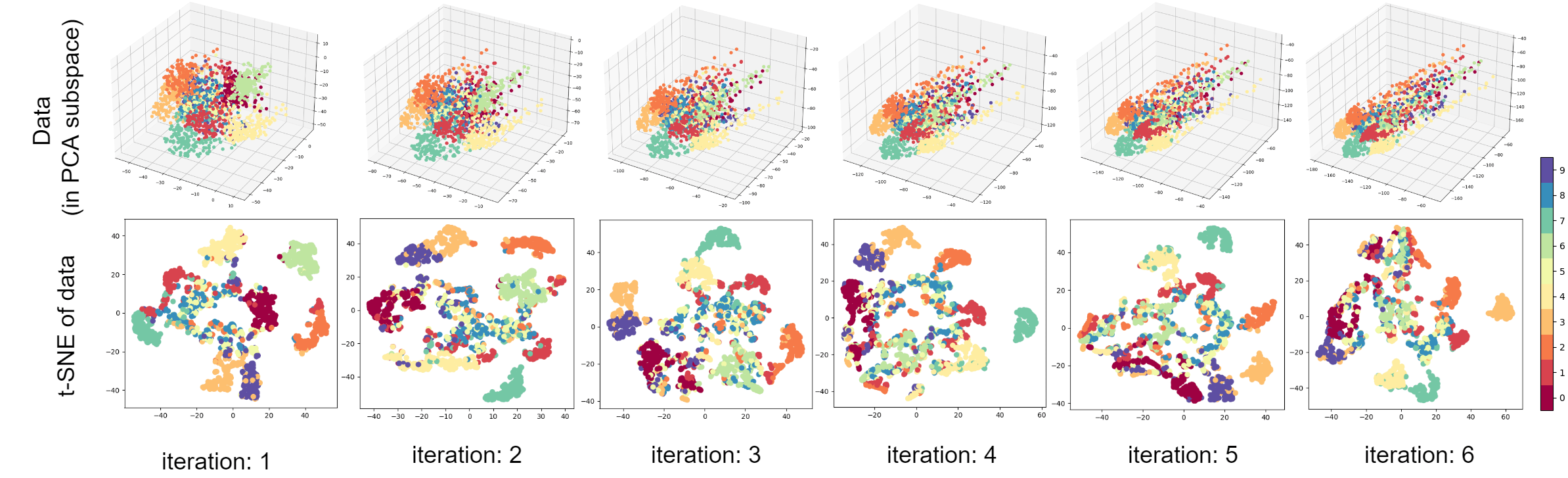}
\caption{Performance of GDR algorithm with relativity gravitation using the Minkowski metric tensor.}
\label{figure_experiments2_Relativity_Minkowski}
\end{figure*}

Figure \ref{figure_experiments2_Relativity_Minkowski} illustrates the performance of GDR algorithm with relativity gravitation using the Minkowski metric tensor and $\alpha_x = \alpha_y = \alpha_z = 0.33$. As can be seen in this figure, the Minkowski metric only performs well up to some point and not as well as the Schwarzschild metric. This is because the Minkowski assumes the space is flat and doe not encounter the effect of mass of $\b{x}_i$ on $\b{x}_j$.

\subsection{Comparison of Weights of Coordinates}

Finally, we compare the performances of GDR algorithm with relativity gravitation approach, having various weights of coordinates. This compares the importance of movements along different spherical coordinates. We compare three extreme cases $\{\alpha_r = 1, \alpha_\theta=0, \alpha_\phi=0\}$, $\{\alpha_r = 0, \alpha_\theta=1, \alpha_\phi=0\}$, and $\{\alpha_r = 0, \alpha_\theta=0, \alpha_\phi=1\}$. The results demonstrate that the radial coordinate has the most important impact in movement. This is expected because movement in $r$ direction brings $\b{x}_j$ to $\b{x}_i$ fastest among the spherical coordinates.
However, as was shown in Fig. \ref{figure_experiments1_Newtonian_and_Relativity}, moving along all spherical coordinates in combination works very well for moving along the geodesic. 

\begin{figure*}[!t]
\centering
\includegraphics[width=\textwidth]{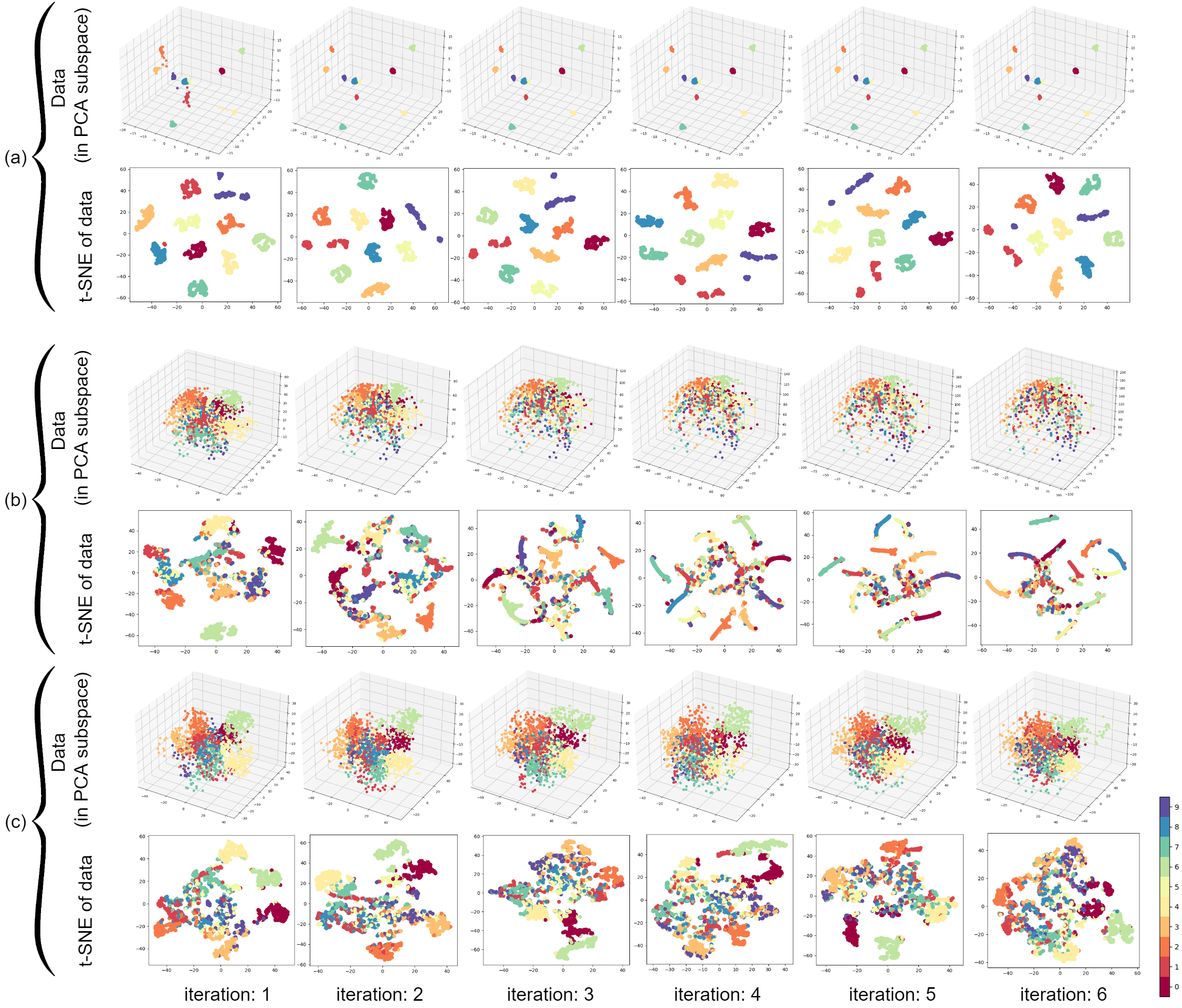}
\caption{Comparison of performance of GDR algorithm with relativity gravitation having (a) $\alpha_r = 1, \alpha_\theta=0, \alpha_\phi=0$ and (b) $\alpha_r = 0, \alpha_\theta=1, \alpha_\phi=0$, and (c) $\alpha_r = 0, \alpha_\theta=0, \alpha_\phi=1$.}
\label{figure_experiments3_Relativity_Minkowski}
\end{figure*}

\section{Conclusion and Future Directions}\label{section_conclusion}

This work proposed the GDR algorithm as a hybrid of physics and machine learning where physics is used in machine learning and not vice versa. Two versions of GDR were proposed, using the Newtonian gravitation and the relativity gravitation. In GDR with relativity, both Schwarzschild and Minkowski metrics were used to have both general relativity and special relativity in GDR. The experiments demonstrated that GDR is effective in separation of classes. 
The proposed algorithms are hybrids of physics and machine learning. The effectiveness of these algorithms were demonstrated by extensive experimentation and comparisons done both quantitatively and qualitatively.  

Several possible directions exist for future work. The proposed GDR algorithm is a supervised method. Developing an unsupervised version of algorithm is a possibility. 
Moreover, the current GDR algorithm does not support out-of-sample (test) data which are not used during training. Out-of-sample extension of GDR is also a possible future direction. 
Another possible future work is consideration of gravitational waves impacting the data points in the GDR algorithm.
Finally, other metric tensors, such as the Friedmann-Lemaitre-Robertson-Walker metric (used in cosmology), the Lemaitre-Tolman metric (used for spherically symmetric dust), and the G{\"o}del metric, can be used in GDR for developing the algorithm.

%%%%%%%%%%%%%%%%%%%%%%%%%%%%%%%%%%%%%%%%%%%%%%%%%%%%%%%%%%%%%%%%%%%%%%%%%%%%%%%%%
\bibliography{ref}
% \bibliographystyle{plainnat}
%%%%%%%%%%%%%%%%%%%%%%%%%%%%%%%%%%%%%%%%%%%%%%%%%%%%%%%%%%%%%%%%%%%%%%%%%%%%%%%%%

% \newpage
% \medskip
% \medskip
% \section*{\centering \huge Appendices }
% \appendix

% %%%%%%%%%%%%%%%%%%%%%%%%%%%%%%%%%%%%%%%%%%%%%%%%%%%%%%%%%%%%%%%%%%%%%
% \section{Preliminaries} \label{app:prelim}

\end{document}